\documentclass{article}
\usepackage{spconf,amsmath,graphicx}
\usepackage{array}
\usepackage{textcomp}

\title{DAPlankton: Benchmark Dataset for Multi-instrument Plankton Recognition via Fine-grained Domain Adaptation}
\name{\begin{minipage}[b]{\linewidth}
   Daniel Batrakhanov$^{\star}$ \qquad Tuomas Eerola$^{\star}$ \qquad Kaisa Kraft$^{\dagger}$ \qquad Lumi Haraguchi$^{\dagger}$ \qquad Lasse Lensu$^{\star}$\\
   \textit{Sanna Suikkanen$^{\ddagger}$ \qquad María Teresa Camarena-Gómez$^{\mathsection}$ \qquad Jukka Sepp\"al\"a$^{\dagger}$ \qquad Heikki K\"alvi\"ainen$^{\star}$}\end{minipage}}
  
  \address{$^{\star}$ Computer Vision and Pattern Recognition Laboratory, LUT University, Finland \\
      $^{\dagger}$ Research Infrastructure, Finnish Environment Institute, Finland \\
      $^{\ddagger}$ Marine and Freshwater Solutions, Finnish Environment Institute, Finland\\
      $^{\mathsection}$ Centro Oceanografico de Malaga, Instituto Español de Oceanografia, Spain}

\begin{document}
%
\maketitle
\begin{abstract}
Plankton recognition provides novel possibilities to study various environmental aspects and an interesting real-world context to develop domain adaptation (DA) methods. Different imaging instruments cause domain shift between datasets hampering the development of general plankton recognition methods. A promising remedy for this is DA allowing to adapt a model trained on one instrument to other instruments. In this paper, we present a new DA dataset called DAPlankton which consists of phytoplankton images obtained with different instruments. Phytoplankton provides a challenging DA problem due to the fine-grained nature of the task and high class imbalance in real-world datasets. DAPlankton consists of two subsets. DAPlankton$_\mathrm{LAB}$ contains images of cultured phytoplankton providing a balanced dataset with minimal label uncertainty. DAPlankton$_\mathrm{SEA}$ consists of images collected from the Baltic Sea providing challenging real-world data with large intra-class variance and class imbalance. We further present a benchmark comparison of three widely used DA methods.

\end{abstract}
\section{INTRODUCTION}
\label{sec:intro}

In various practical applications of image recognition (e.g., environmental, industrial, and medical imaging) the data consist of a large pool of relatively small in-house datasets. There are large domain shifts between these datasets caused by, for example, different imaging setups and environments. Despite being able to provide human-level accuracy in restricted image classification tasks, the learned image representations (e.g., CNNs) are known to be weak at generalizing beyond the domain they were trained on~\cite{gulrajani2020}. A remedy for this is the unsupervised domain adaptation that allows applying a method trained on one dataset efficiently to new datasets from a different domain without having labeled training data in the new domain.
\begin{figure}[t]
\begin{center}
   \includegraphics[width=0.8\linewidth]{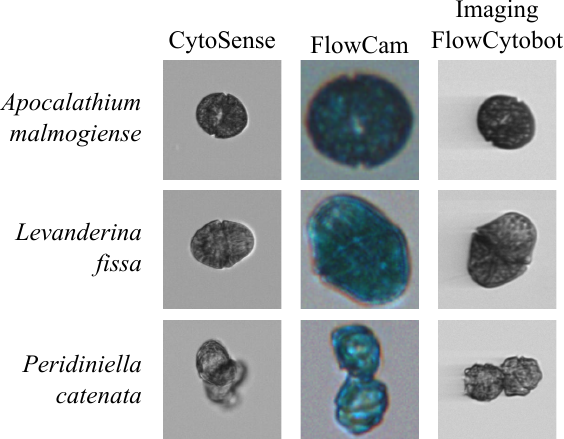}
\end{center}
   \caption{Phytoplankton images from DAPlankton$_{LAB}$. Each column contains a different imaging instrument (domain) and each row a different phytoplankton species.}
\label{fig:intro}
\end{figure}

Automatic plankton recognition has the potential to revolutionize plankton research providing new possibilities to study plankton populations and various environmental aspects such as marine food webs and CO$_2$ exchange between the sea and atmosphere. At the same time, plankton recognition provides a good context for developing domain adaptation methods for image recognition. Different imaging instruments and local differences in plankton species composition have caused a large pool of varying plankton image datasets. High-quality labeled data relies mostly on trained taxonomists, which heavily limits the amount of training data in datasets lowering the recognition accuracy of the models. More importantly, new datasets are constantly being collected and it is not feasible to label training data and to train a model from scratch for each of them separately. Methods are needed to adapt plankton recognition for new datasets with minimal supervision.

In this paper, we introduce a new dataset called DAPlankton for developing and benchmarking domain adaptation methods for image recognition. The data consists of phytoplankton images captured using different imaging instruments resulting in domain shift. This provides a challenging and applicationally relevant dataset that is naturally multi-domain unlike some existing domain adaptation benchmarks that rely on more artificial settings. Moreover, phytoplankton image data introduces additional challenges including the fine-grained nature of the recognition problem and highly unbalanced class distribution.

The data is divided into two subsets: DAPlankton$_\mathrm{LAB}$ and DAPlankton$_\mathrm{SEA}$. DAPlankton$_\mathrm{LAB}$ consists of images captured from multiple mono-specific phytoplankton cultures, which were analysed using three different imaging instruments: Imaging FlowCytobot (IFCB)~\cite{olson2007submersible}, CytoSense (CS)~\cite{dubelaar1999design}, and FlowCam (FC)~\cite{sieracki1998imaging} each producing cropped images with one plankton particle in each (see Fig.~\ref{fig:intro}). An expert further verified the class of each image, ensuring that there was no cross-contamination between different cultures. This process resulted in a balanced dataset with negligible label uncertainty. DAPlankton$_\mathrm{SEA}$ consists of images captured from water samples collected from the Baltic Sea using two different imaging instruments: IFCB and CS. Each image was manually labeled by an expert. DAPlankton$_\mathrm{SEA}$ provides a realistic and more challenging dataset with a large class imbalance and natural intra-class variance. The DAPlankton dataset has been made publicly available for research purposes.\footnote{https://doi.org/10.23729/32583bd0-38cd-4532-a8d6-fc9dc5967dce}

We provide an evaluation protocol, as well as a preliminary benchmark study on three widely used domain adaptation methods. We focus on unsupervised closed-set domain adaptation, that is, each domain (imaging instrument) contains the same set of classes, and no labels from the target domain are provided for the adaptation method. The results show that commonly used domain adaptation methods are not able to tackle the fine-grained nature of the plankton recognition task. This calls for novel methods that can simultaneously address the domain shift between different imaging instruments, large intra-class variation, and small inter-class variation.

\section{RELATED WORK}
\subsection{Domain adaptation}

Domain adaptation (DA) represents a particular case of transfer learning in which a distribution shift in data between the source and target domains occurs, but the initial task is shared. 
%
Due to the prevalence of domain shift in various real-world recognition problems, a large number of DA methods have been proposed. From the methodological perspective, DA approaches can be categorized into five main categories: 1)~instance re-weighting adaptation, 2)~feature adaptation, 3)~classifier adaptation, 4)~deep network adaptation, and 5)~adversarial adaptation~\cite{zhang2022transfer}.  In instance re-weighting, weights of the source data samples are weighted to better match the distribution of the target data (e.g.,~\cite{yan2017mind}). In feature adaptation, the goal is to learn common feature representations that have similar source and target distributions (e.g.,~\cite{sun2016}). Classifier adaptation aims to learn a generalized classifier, for example, using dynamic distribution adaptation~\cite{wang2020transfer}. Deep network adaptation includes methods based on maximum mean discrepancy (e.g.,~\cite{long2017deep}) and autoencoders (e.g.,~\cite{ghifary2016deep}) aiming to learn a domain-independent encoder in an unsupervised manner. Finally, methods based on adversarial adaptation follow a similar principle to generative adversarial networks (GANs). In a typical method, a domain discriminator is a model that is trained to distinguish between the source and target domains, and the classification model learns transferable feature representations that the domain discriminator cannot distinguish (e.g.,~\cite{long2018conditional}). An alternative approach is to use GAN-based style transfer models to perform the adaptation at the pixel level (see e.g.,~\cite{murez2018image}).



\subsection{Fine-grained domain adaptation}
The fine-grained recognition task further complicates the domain adaptation. A handful of methods have been proposed to simultaneously address the large inter-domain variations and small inter-class variations. Most of these utilize manual annotation of attributes or part landmarks that makes the method very application-specific and unable to generalize to new recognition problems (see e.g.~\cite{gebru2017fine}). More general methods include domain-specific transfer learning~\cite{cui2018} and progressive adversarial networks~\cite{wang2020}.

\subsection{Plankton recognition}
The research on automatic plankton image recognition has matured from early works based on hand-engineered image features \cite{sosik2007automated} to feature learning-based approaches utilizing deep learning and especially CNNs~\cite{orenstein2017transfer,lumini2019deep,burevs2021plankton,kraft2022towards,badreldeen2022open} and vision transformers~\cite{maracani2023}. Various custom methods and modifications to general-purpose techniques have been proposed to address the special characteristics of plankton image data. For an extensive survey on automatic plankton recognition, see~\cite{eerola2023survey}.

Different imaging instruments cause a domain shift between plankton datasets. This is why most automatic plankton recognition solutions focus on just one imaging instrument. While domain adaptation has not been widely studied on plankton recognition, there have been works where multiple plankton image datasets have been utilized to solve the recognition task. 
For example, transfer learning and fine-tuning have been utilized as approaches against the differences in datasets~\cite{orenstein2017transfer,guo2022cdfm}.

Plonus et al.~\cite{plonus2021automatic} suggest using capsule neural networks combined with probability filters to address the dataset shift caused by different plankton imaging instruments. The idea of capsule neural networks is to form groups of neurons (capsules) that learn the specific properties of the object in the image. Capsule neural networks can be assumed to be less sensitive to the changes in the field conditions and therefore able to adapt to different data distributions. 

\subsection{Existing benchmark datasets}

Multiple datasets for domain adaptation exist. These include Office-Home~\cite{venkateswara2017deep}, Office-31~\cite{saenko2010adapting}, VisDA2017~\cite{peng2017visda}, and DomainNet~\cite{peng2019moment}.
One notable drawback of many existing datasets is that they are based on rather artificial settings with domains such as drawings and synthetic images in addition to camera images. 
Only few datasets for fine-grained domain adaptation exist: CompCars~\cite{yang2015large} (78k images from two domains) and CUB-Paintings~\cite{wang2020} (15k images from two domains).
In this paper, a fine-grained image recognition dataset that is naturally multi-domain and provides an environmentally relevant application is provided to test domain adaptation methods.

\section{DATA}
The data consists of two subsets: DAPlankton$_\mathrm{LAB}$ and DAPlankton$_\mathrm{SEA}$. DAPlankton$_\mathrm{LAB}$ consists of mono-specific phytoplankton cultures and DAPlankton$_\mathrm{SEA}$ consists of natural water samples obtained from the Baltic Sea.

\subsection{Data collection and manual labeling}
\subsubsection{DAPlankton$_\mathrm{LAB}$}
Mono-specific phytoplankton cultures, with organisms isolated from the Baltic Sea (FINMARI Culture Collection/Syke Marine Research Laboratory and Tvärminne Zoological Station) and Danish coastal waters, were selected to represent different morphologies and taxonomical groups. Cultures are kept under stable conditions, optimized for the organisms growth. \textit{Aphanizomenon flosaquae} (Cyanophyta), \textit{Chrysotila roscoffensis} (Haptophyta), \textit{Kryptoperidinium folliaceum} (Dinophyta), \textit{Levanderina fissa} (Dinophyta) and \textit{Pseudopedinella} sp. (Ochrophyta) are kept in 6\textperthousand~ salinity and 16 \textdegree C. \textit{Apocalathium malmogiense} (Dinophyta), \textit{Diatoma tenuis} (Bacillariophyta), \textit{Gymnodinium corollarium} (Dinophyta), \textit{Melosira arctica} (Bacillariophyta), \textit{Peridiniella catenata} (Dinophyta) and \textit{Rhinomonas nottbeckii} (Cryptophyta) are kept in 6 \textperthousand~ salinity and 4 \textdegree C. \textit{Nephroselmis pyriformis} (Chlorophyta), \textit{Rhodomonas salina} (Cryptophyta), \textit{Teleaulax acuta} (Cryptophyta) and \textit{Tetraselmis} sp. (Chlorophyta) are kept in 12 \textperthousand~ salinity and 10 \textdegree C. 
All cultures are maintained at a 14:10 light:dark cycle using $f/2$ medium~\cite{guillard1962studies}, prepared with filtered and autoclaved Baltic Sea water (for cultures grown in 6 \textperthousand~ salinity) or Baltic Sea water with adjusted salinity using artificial sea water (for cultures grown in 12 \textperthousand~ salinity). 
Each culture was analysed individually with CytoSense, FlowCam and IFCB. CytoSense was equipped with two lasers (488 and 594 $nm$), sensors for light scatter and fluorescence and a high resolution camera (3.6 pixels/$\mu m$). Samples were run using a chlorophyll a fluorescence trigger and a pump speed of 1 $\mu l/s$. FlowCam analyses were conducted using a colored camera, flowcell FC100 and a 10$\times$ objective, with images collected in auto-trigger mode. IFCB samples were run using a chlorophyll a fluorescence trigger. 
All images were manually checked to exclude poor-quality images and potential cross-contamination between cultures and assigned to the taxonomical classes based on the strain name (Table~\ref{tab:data_lab}).

\subsubsection{DAPlankton$_\mathrm{SEA}$}
CytoSense images from natural samples from the Baltic Sea were collected during summer 2020 (July and August) at Ut\"o Atmospheric and Marine Research Station ($59^\circ$$46.84'$ N $21^\circ$$22.13'$ E) -- see \cite{kraft2021first} for further details. The instrument used had the same laser and sensors as described above. To optimize different phytoplankton size classes, two runs were set with different chlorophyll a fluorescence levels (30 and 80 $mV$) and sampling pump speeds (2 and 5 $\mu l/s$), targeting smaller and larger organisms, respectively. The IFCB data collection is described in \cite{kraft2021first} and \cite{kraft2022towards}. In short, the instrument was used in continuous mode with a new sample starting every approx. 23 minutes. A chlorophyll a trigger was used and there was a 150 µm mesh at the instrument inlet to prevent it from clogging. Both IFCB and CS images were manually sorted by experts, and good-quality images were assigned to different classes. CS classes followed the ones used in SYKE-plankton\_IFCB\_2022~\footnote{http://doi.org/10.23728/b2share.abf913e5a6ad47e6baa273ae0ed6617a}. 

\subsection{Data composition}

\subsubsection{DAPlankton$_\mathrm{LAB}$}

\begin{figure}[t]
\begin{center}
   \hspace{0.23cm} CS \hspace{2cm} FC \hspace{1.8cm} IFCB
   \includegraphics[width=0.9\linewidth]{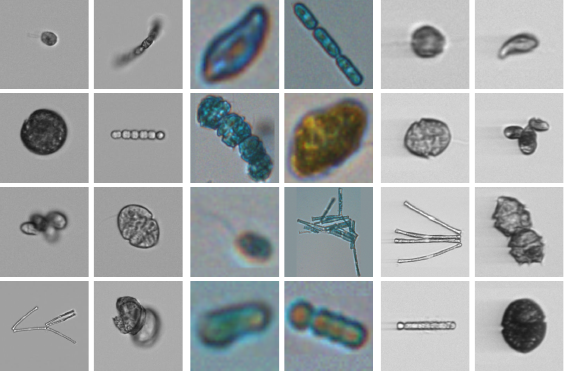}\\
   \caption{Example images from DAPlankton$_\mathrm{LAB}$. Domains from left to right: CytoSense (CS), FlowCam (FC), and Imaging FlowCytobot (IFCB).}
\label{fig:examples_lab}
\end{center}
\end{figure}

DAPlankton$_\mathrm{LAB}$ contains, in total, 47~471 images from 15 phytoplankton species and 3 different domains (imaging instruments). The number of images per class-domain combination varies between 286 and 2618 providing a reasonably balanced dataset. Table~\ref{tab:data_lab} shows the classes and the numbers of images for each imaging instrument. Example images from DAPlankton$_\mathrm{LAB}$ are shown in Fig.~\ref{fig:examples_lab}.

\begin{table}
\begin{center}
\caption{Composition of DAPlankton$_\mathrm{LAB}$.\label{tab:data_lab}}
\begin{tabular}{|l|c|c|c|}
\hline
Class & CS & FC & IFCB \\
\hline\hline
\textit{Aphanizomenon flosaquae} & 926 & 1138 & 1088 \\
\textit{Apocalathium malmogiense} & 936 & 1647 & 1003 \\
\textit{Chrysotila roscoffensis} & 701 & 453 & 1064 \\
\textit{Diatoma tenuis} & 862 & 939 & 1376 \\
\textit{Gymnodinium corollarium} & 818 & 2618 & 1072 \\
\textit{Kryptoperidium foliaceum} & 1285 & 1030 & 1075 \\
\textit{Levanderina fissa} & 870 & 867 & 1001 \\
\textit{Melosira arctica} & 608 & 826 & 1002 \\
\textit{Nephroselmis pyriformis} & 826 & 286 & 1326 \\
\textit{Peridiniella catenata} & 757 & 585 & 1250 \\
\textit{Pseudopedinella} sp. & 1088 & 1849 & 1018 \\
\textit{Rhinomonas nottbecki} & 1037 & 1394 & 1001 \\
\textit{Rhodomonas salina} & 877 & 2232 & 1149 \\
\textit{Teleaulax acuta} & 762 & 486 & 1011 \\
\textit{Tetraselmis} sp. & 834 & 1458 & 1040 \\
\hline\hline
Total & 13187 & 17808 & 16476 \\
\hline
\end{tabular}
\end{center}
\end{table}

\subsubsection{DAPlankton$_\mathrm{SEA}$}

\begin{figure}[htb]

\begin{minipage}[b]{0.1\linewidth}
 \centering
 CS\\
 \vspace{2.2cm}
 IFCB\\
 \vspace{1.8cm}
\end{minipage}
\begin{minipage}[b]{0.29\linewidth}
 \centering
 \centerline{\includegraphics[width=2.42cm]{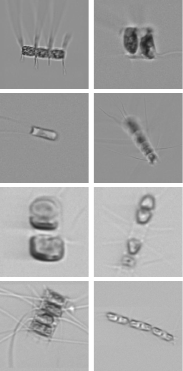}}
 \centerline{(a)}\medskip
\end{minipage}
\begin{minipage}[b]{0.29\linewidth}
 \centering
 \centerline{\includegraphics[width=2.42cm]{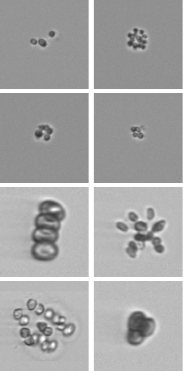}}
 \centerline{(b)}\medskip
\end{minipage}
\begin{minipage}[b]{0.29\linewidth}
 \centering
 \centerline{\includegraphics[width=2.42cm]{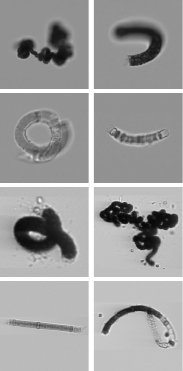}}
 \centerline{(c)}\medskip
\end{minipage}
\caption{Example images from the DAPlankton$_\mathrm{SEA}$ dataset. Notice the large intra-class variation: (a) \textit{Chaetoceros} sp.; (b) Chlorococcales; (c)~\textit{Nodularia spumigena.}}
\label{fig:examples_sea}

\end{figure}

DAPlankton$_\mathrm{SEA}$ contains, in total, 64~453 images from 31 plankton classes and 2 different domains. Table~\ref{tab:data_sea} shows the classes and the numbers of images for each imaging instrument. As can be seen, the class distribution is considerably less balanced (the number of images per class-domain combination varies between 5 and 12~280) as the class sizes correspond to the true rarity of the plankton class. Moreover, the intra-class variance is larger as can be seen in Fig.~\ref{fig:examples_sea}.

\begin{table}[t]
\begin{center}
\caption{Composition of DAPlankton$_\mathrm{SEA}$.\label{tab:data_sea}}
\begin{tabular}{|l|c|c|}
\hline
Class & CS & IFCB \\
\hline\hline
\textit{Aphanizomenon flosaquae} & 2027 & 6989 \\
Centrales sp. & 33 & 480 \\
\textit{Chaetoceros} sp. & 43 & 1382 \\
\textit{Chaetoceros} sp. single & 5 & 213 \\
Chlorococcales & 17 & 95 \\
Chroococcales & 938 & 142 \\
Ciliata & 174 & 243 \\
Cryptomonadales & 177 & 713 \\
Cryptophyceae Teleaulax & 5443 & 6830 \\
\textit{Cyclotella choctawhatcheeana} & 33 & 102 \\
Dinophyceae & 275 & 1433 \\
\textit{Dinophysis acuminata} & 7 & 217 \\
\textit{Dolichospermum Anabaenopsis} & 109 & 12280 \\
\textit{Dolichospermum Anabaenopsis} coiled & 165 & 2504 \\
Euglenophyceae & 67 & 102 \\
\textit{Eutreptiella} sp. & 1027 & 2247 \\
Gymnodiniales & 7 & 69 \\
\textit{Gymnodinium} like & 44 & 158 \\
\textit{Heterocapsa rotundata} & 330 & 614 \\
\textit{Heterocapsa triquetra} & 96 & 3276 \\
Heterocyte & 240 & 263 \\
\textit{Katablepharis remigera} & 14 & 54 \\
\textit{Mesodinium rubrum} & 27 & 1132 \\
\textit{Monoraphidium contortum} & 352 & 327 \\
\textit{Nitzschia paleacea} & 11 & 65 \\
\textit{Nodularia spumigena} & 66 & 169 \\
\textit{Oocystis} sp. & 100 & 842 \\
\textit{Pseudopedinella} sp. & 57 & 379 \\
\textit{Pyramimonas} sp. & 851 & 1224 \\
\textit{Skeletonema marinoi} & 35 & 4128 \\
\textit{Snowella Woronichinia} & 61 & 2950 \\
\hline\hline
Total & 12831 & 51622 \\
\hline
\end{tabular}
\end{center}
\vspace{-15pt}
\end{table}

\section{BASELINE METHODS}

Three commonly used domain adaptation methods are included as baseline methods: Deep Correlation Alignment (Deep CORAL)~\cite{sun2016}, Conditional Adversarial Domain Adaptation (CDAN)~\cite{long2018conditional}, and Manifold Embedded Distribution Alignment (Deep MEDA)~\cite{wang2020transfer}.
These were selected to have representative examples from the three main approaches for domain adaptation: feature adaptation (Deep CORAL), adversarial learning (CDAN), and classifier adaptation (MEDA).


\subsection{Deep CORAL}

The main idea behind Deep CORAL~\cite{sun2016} is to learn domain invariant image features. This is achieved by using CORAL loss that minimizes the difference in second-order statistic (convariances) between source and target feature distributions, and this way, reduces the domain gap.

\subsection{CDAN}

CDAN~\cite{long2018conditional} is an example of adversarial learning-based domain adaptation, where a domain discriminator is trained to distinguish between the source and target domains, and the classification model learns transferable feature representations that the domain discriminator cannot distinguish. CDAN further conditions the domain discriminator on the cross-covariance of domain-specific feature representations and classifier predictions.

\subsection{Deep MEDA}

MEDA~\cite{wang2020transfer} relies on geodesic flow kernel for bridging the domains in Grassmann manifold, and maximum mean discrepancy to measure a divergence between domain in the transformed feature subspaces. The feature representations embedded in Grassmannians contribute to more accurate reflection of domain properties and structure preventing feature distortion. The deep version of MEDA relies on CNN features.

\begin{table*}[t]
\begin{center}
\caption{Results on DAPlankton$_\mathrm{LAB}$.\label{tab:results_lab}}
\begin{small}
\begin{tabular}{|l|l|c|c|c|c|c|c|c|}
\hline
Method & Backbone & IFCB $\rightarrow$ CS & IFCB $\rightarrow$ FC & CS $\rightarrow$ IFCB & CS $\rightarrow$ FC & FC $\rightarrow$ IFCB & FC $\rightarrow$ CS & Average \\
\hline \hline
No adaptation & AlexNet & 64.5 $\pm$ 0.8 & 31.7 $\pm$ 0.9 & 51.6 $\pm$ 0.5 & 19.0 $\pm$ 1.4 & 38.5 $\pm$ 0.3 & 25.9 $\pm$ 1.4 & 38.5 $\pm$ 0.6 \\
 & ResNet & 40.1 $\pm$ 1.1 & 13.8 $\pm$ 1.0 & 48.5 $\pm$ 2.3 & 6.3 $\pm$ 1.7 & 39.3 $\pm$ 2.4 & 25.4 $\pm$ 2.0 & 28.9 $\pm$ 0.5 \\
Deep CORAL & AlexNet & 69.8 $\pm$ 0.4 & 38.5 $\pm$ 2.2 & 69.7 $\pm$ 0.6 & 25.3 $\pm$ 2.1 & 49.9 $\pm$ 2.0 & 36.1 $\pm$ 0.5 & 48.2 $\pm$ 0.5 \\
 & ResNet & 63.1 $\pm$ 0.5 & 22.9 $\pm$ 1.5 & 69.1 $\pm$ 1.6 & 19.1 $\pm$ 1.8 & 38.9 $\pm$ 1.8 & 27.7 $\pm$ 4.0 & 40.1 $\pm$ 0.4 \\
CDAN & AlexNet & \textbf{84.7 $\pm$ 1.6} & 29.6 $\pm$ 1.5 & 86.9 $\pm$ 0.9 & 17.9 $\pm$ 2.0 & 44.7 $\pm$ 3.7 & 33.6 $\pm$ 0.9 & 49.5 $\pm$ 1.1 \\
 & ResNet & 44.5 $\pm$ 0.7 & 20.4 $\pm$ 0.9 & 50.4 $\pm$ 0.6 & 16.9 $\pm$ 1.2 & 40.3 $\pm$ 0.9 & 17.5 $\pm$ 2.0 & 31.7 $\pm$ 0.5 \\
Deep MEDA & AlexNet & 81.3 $\pm$ 0.8 & 54.8 $\pm$ 1.8 & \textbf{87.7 $\pm$ 1.2} & \textbf{47.4 $\pm$ 4.3} & 71.3 $\pm$ 2.1 & 45.5 $\pm$ 3.1 & 64.7 $\pm$ 1.2 \\
 & ResNet & 81.1 $\pm$ 1.3 & \textbf{55.5 $\pm$ 3.1} & 87.6 $\pm$ 0.9 & 43.1 $\pm$ 5.4 & \textbf{73.8 $\pm$ 5.7} & \textbf{60.9 $\pm$ 2.6} & \textbf{67.0 $\pm$ 1.6} \\
\hline
\end{tabular}
\end{small}
\end{center}
\vspace{-10pt}
\end{table*}

\begin{table*}
\begin{center}
\caption{Results on DAPlankton$_\mathrm{SEA}$.\label{tab:results_sea}}
\begin{tabular}{|l|l|c|c|c|}
\hline
Method & Backbone & IFCB $\rightarrow$ CS & CS $\rightarrow$ IFCB & Average \\
\hline\hline
No adaptation & AlexNet & 51.1 $\pm$ 2.4 & 61.7 $\pm$ 1.0 & 56.4 $\pm$ 1.6 \\
 & ResNet & 32.5 $\pm$ 3.3 & 50.3 $\pm$ 3.3 & 41.4 $\pm$ 1.4 \\
Deep CORAL & AlexNet & 42.9 $\pm$ 1.2 & 53.8 $\pm$ 2.7 & 48.3 $\pm$ 1.1 \\
 & ResNet & 41.8 $\pm$ 1.6 & 46.8 $\pm$ 1.0 & 44.3 $\pm$ 0.7 \\
CDAN & AlexNet & \textbf{76.3 $\pm$ 0.9} & \textbf{64.8 $\pm$ 1.4} & \textbf{70.6 $\pm$ 1.0} \\
 & ResNet & 25.9 $\pm$ 3.3 & 62.3 $\pm$ 2.4 & 44.1 $\pm$ 2.4 \\
Deep MEDA & AlexNet & 51.9 $\pm$ 0.5 & 57.3 $\pm$ 0.9 & 54.6 $\pm$ 0.3 \\
 & ResNet & 53.4 $\pm$ 0.6 & 59.9 $\pm$ 0.3 & 56.6 $\pm$ 0.2 \\
\hline
\end{tabular}
\end{center}
\vspace{-10pt}
\end{table*}

\section{EVALUATION PROTOCOL}

For evaluation, each subset-domain combination is divided into training, and test set using the following split: 80\% for training and 20\% for testing. We consider unsupervised domain adaptation, i.e. the domain adaptation model does not have access to any labels from the target domain during the training phase. For each experiment, the evaluated model is trained using all the data in the source domain with ground truth labels and training subset of the target domain without ground truth labels. The evaluation is carried out using the test subset in the target domain.

The recognition results are evaluated using classification accuracy defined as follows:
\begin{equation}
\mathrm{accuracy} = \frac{|\{\mathbf{x}|\mathbf{x}_i \in \mathcal{D}_{\mathrm{test}} \wedge f(\mathbf{x}_i) = y_i\}|}{|\mathcal{D}_{\mathrm{test}}|},
\end{equation}
where $\mathbf{x}_i$ is the input instance (image), $\mathcal{D}_{\mathrm{test}}$ is the test set of the target domain, $f(\mathbf{x}_i)$ is the model prediction and $y_i$ is the corresponding ground truth label. 

The accuracy is computed for all tasks, that is source domain and target domain combinations denoted as Source $\rightarrow$ Target. DAPlankton$_\mathrm{LAB}$ has 3 domains resulting in $3 \times 2 = 6$ tasks. 
DAPlankton$_\mathrm{SEA}$ has 2 tasks: IFCB $\rightarrow$ CS and CS $\rightarrow$ IFCB. The evaluation is repeated for each model and task 5 times with different training set/test set splits and the average accuracies and standard deviations are calculated.



\section{RESULTS AND DISCUSSION}

The three selected domain adaptation methods were applied with two backbone architectures: AlexNet and ResNet-18. These were selected since AlexNet and ResNet are the backbones utilized in the original implementations of the selected methods. Moreover, ResNet-18 has been shown to achieve a high plankton recognition accuracy~\cite{kraft2022towards}. The need for domain adaptation was further demonstrated by training the model on the source dataset and applied to target dataset as-is (no adaptation).
The results are shown in Tables~\ref{tab:results_lab} and~\ref{tab:results_sea}. 

While the domain adaptation methods outperform the recognition models applied without adaptation, the accuracies are still rather low.
This implies that these commonly used unsupervised domain adaptation methods are not able to tackle the fine-grained recognition task, i.e. the small inter-class variation and large intra-class variation. Adaptation between IFCB and CS results in higher accuracy than between FC and the other modalities. This is expected as the domain gap between the IFCB and CS is considerably smaller (see Fig.~\ref{fig:intro}). 

CDAN produced the best results when adapting between IFCB and CS in both DAPlankton$_\mathrm{LAB}$ and DAPlankton$_\mathrm{SEA}$, however, the accuracy drops significantly when the domain gap increases (see e.g. CS $\rightarrow$ FC). This is most likely due to the domain discriminator module and the adversarial learning. A known problem with adversarial-based alignment methods is that the global alignment strategy does not take into account the class-level distributions~\cite{luo2019taking}. Because of this, the classes that are well aligned between the source and target domain can be incorrectly mapped. This effect is emphasized when the domain gap is large and the inter-class variation is small.

Overall, the best method for DAPlankton$_\mathrm{LAB}$ was Deep MEDA providing similar accuracy to CDAN when the domain gap is small and outperforming the other methods when the domain gap is large (adapting from and to FC). However, a clear drop in accuracy compared to CDAN is observed when applying the methods to DAPlankton$_\mathrm{SEA}$. This is likely due to the larger intra-class variance and its effect on the dynamic distribution alignment that Deep MEDA utilizes.

A surprising result is that the AlexNet backbone outperforms ResNet-18 for most methods and tasks. It is possible that the deeper ResNet model learns more domain-specific image features while the shallower AlexNet learns more generalizable features. Also, it is good to note that Deep CORAL was originally developed for AlexNet, and therefore, some design choices might be more suited for that backbone architecture.

\section{CONCLUSIONS}
This paper describes a novel benchmark dataset for fine-grained domain adaptation called DAPlankton. The dataset consists of over 110k expert-labeled phytoplankton images captured with different imaging instruments representing the domains. 
Plankton provides a naturally multi-domain and challenging dataset that is characterized by the fine-grained nature of the recognition problem. The dataset was designed to develop and evaluate unsupervised domain adaptation methods, i.e. given a source dataset with expert labels and a target dataset without labels, the task is to adapt the recognition model trained on the source dataset to the target dataset. We further provided an evaluation protocol and performed a preliminary benchmark study on three commonly used domain adaptation methods. The results suggest that these methods are not capable of tackling the domain shift between different imaging instruments, large intra-class variation, and small inter-class variation. This calls for novel methods for the fine-grained domain adaptation.

\section{ACKNOWLEDGMENTS}

The research was carried out in the FASTVISION and FASTVISION-plus projects funded by the Academy of Finland (Decision numbers 321980, 321991, 339612, and 339355). Lumi Haraguchi was supported by OBAMA-NEXT (grant agreement no. 101081642), funded by the European Union under the Horizon Europe program.

\bibliographystyle{IEEEbib}
\bibliography{references}

\end{document}